\documentclass[11pt]{article}

\usepackage{enumitem}
\usepackage{algorithmic}
\usepackage{algorithm}
\usepackage{xcolor}
\usepackage{xspace}
\usepackage{graphicx}
\usepackage{footnote}
\usepackage{hyperref}
\usepackage{multirow}
\usepackage{subfig}
\usepackage{enumitem}
\usepackage{comment}
\usepackage{amsfonts}
\usepackage{amsmath}

\newcommand{\etal}{{et al}.\@ }
\newcommand{\eg}{{e.g}.\@ }
\newcommand{\ie}{{i.e}.\@ }

\begin{document}

\title{Machine learning and high dimensional vector search}

\date{}

\author{Matthijs Douze, Meta FAIR}

\maketitle

\begin{abstract}
    Machine learning and vector search are two research topics that developed in parallel in nearby communities. 
    However, unlike many other fields related to big data, machine learning has not significantly impacted vector search. 
    In this opinion paper we attempt to explain this oddity. 
    Along the way, we wander over the numerous bridges between the two fields. 
\end{abstract}

\renewcommand\thesection{\arabic{section}}
\setcounter{section}{0}
\setcounter{figure}{0}
\setcounter{table}{0}

\section{Introduction}

Most high-dimensional vector search methods are based on statistical tools, signal processing approaches or graph traversal algorithms.
Statistical tools include random projections~\cite{datar2004locality}, dimensionality reduction (PCA and the SVD). 
Signal processing is employed primarily to compress vectors with quantization~\cite{jegou2010product,babenko2014additive,gao2024rabitq} 
Most recent indexing methods are rely on graphs~\cite{malkov2018efficient,subramanya2019diskann,azizi2025graph,chen2024roargraph} that are built with graph traversal heuristics.

Vector search (VS) is used in machine learning (ML) for training data deduplication~\cite{oquab2024dinov2learningrobustvisual} and searching ML embeddings~\cite{izacard2020leveraging,babenko2014neural}. 
Therefore, there are many research teams around the world that are competent in both fields. 

In their seminal work \emph{The case for learned indexing structures}~\cite{kraska2018case}, Kraska \etal introduced a series of ML based algorithms to speed up classical indexing structures like hash tables and B-trees. 
These structures are building blocks of databases, \eg a B-tree can be seen as a one dimensional VS index.
They hoped to open ``an entirely new research direction for a decades old field''. 

However, 7 years later, it is striking that ML has had very little impact on VS. 
No ML based method appears in the current VS benchmarks at any scale ~\cite{aumuller2020ann,bigann21,bigann23}.
The most advanced ML tool that is widely used for VS is the k-means clustering algorithm. 
There is no deep learning, no use of high-capacity networks. 

In the following, we attempt to explain why this is the case. 
We work out a typical use case of VS (Section~\ref{sec:IR}), then expose the fundamental limits of ML for VS (Section~\ref{sec:ML4VS}); 
in Section~\ref{sec:VS4ML}, we review interesting applications in the other direction: VS for ML.

\section{Information retrieval with machine learning}
\label{sec:IR}

Let's start from a typical application of vector search: image retrieval.
This is originally a classical ML problem, and we show how it converts to VS.

We manage a collection of $N$ images $C=\{x_1, \dots, x_N\}$, where $N$ is large, and possibly growing. 
From a given query image $q$, the objective is to find images in $C$ that are relevant to it, for example because they represent the same object.

\paragraph{$N$-way classifier.}
The most straightforward ML approach for this is to train a $N$-way classifier $f_1(q)\in[0,1]^N$ that outputs a scalar score close to 1 for the images that are relevant and close to 0 for the others. 
The best matching item from the collection can be found without even accessing the images after the training phase: 
\begin{equation}
    I_1 = \mathrm{argmax}~ f_1(q).
\end{equation}

The model could be any standard image classification model, convolutional or transformer-based~\cite{he2016deep,dosovitskiy2021imageworth16x16words}.
Note that processing images is inherently slow, since the raw pixels must be converted into semantically relevant information, which requires several neural net layers. 
More importantly, the model's size needs to be proportional to the collection size $N$, and it is impossible to add or remove images from the collection after the  model was trained. %

\paragraph{Pairwise comparison.}

Another approach is to design a function $f_2$ that, given $(x,x')$ returns a score that is high if $x'$ relevant to $x$ and low otherwise. 
At search time, $f_2$ is used to compare the query $q$ with all the images from the collection (which must be accessible): 
\begin{equation}
    I_2 = \underset{i=1,\dots,N}{\mathrm{argmax}}~ f_2(q, x_i).
\end{equation}

The usual learning machinery can be deployed to train this function: given a training set of matching and non-matching images, the model is optimized to return 0 or 1 with, for example, a binary cross-entropy loss. 

This resolves the problem of the model size, that remains fixed, and adding/removing from the collection is painless. 
The issue with this approach is that the model $f_2$ is still expensive to evaluate, and querying one image requires $N$ evaluations (forward passes) of $f_2$.
Therefore it is not tractable beyond a few hundred images. 

\paragraph{Embeddings.}

A way to avoid this is to force $f_2$ to decompose as: 
\begin{equation}
    f_3(x, x') = S(E(x), E(x')),
\end{equation}
where $E$, the embedder (or feature extractor) is a function that computes a vector in $d$ dimensions from an image, and $S: \mathbb{R}^d \times \mathbb{R}^d \rightarrow \mathbb{R}$ is a similarity function between embeddings. 
This is akin to kernel methods~\cite[chapter 7]{bach2024learning}, and sometimes called a ``two-tower'' network.

The feature extraction function $E$ can be costly to evaluate, but it is performed beforehand, when the image is added to the collection: $(e_1, \dots, e_N) = (E(x_1), \dots, E(x_N))$ is precomputed.
At search time, we perform only simple vector comparisons with all $N$:
\begin{equation}
    I_3 = \underset{i=1,\dots,N}{\mathrm{argmax}}~ S(E(q), e_i)
\label{eq:embeddingsearch}
\end{equation}

From a ML point of view, since we impose a constraint on the function $f_3$, it is bound to be \emph{less accurate} than the function $f_2$, that performs a direct comparison between items: there is an information bottleneck. 

The functions $S$ and $E$ are bound by an \emph{embedding contract}~\cite{douze2024faiss}, that states that $E$ should generate embeddings such that the comparison with $S$ is semantically meaningful. 

\paragraph{Embedding extraction.}

There is a vast literature on embedding extraction for images. 
It started in the pre-deep era~\cite{jegou2010aggregating,perronnin2010large}. 
Later, image embedding models were first sampled from some layer of a classification model~\cite{babenko2014neural}, then trained specifically using a clustering supervision~\cite{caron2018deep}, or variants of contrastive learning~\cite{deng2019arcface,caron2021emerging,chen2020simple, jaiswal2020survey}. 
The embeddings can be specialized for different granularities of image matching: from class-level via instance-level to copy-level~\cite{berman2019multigrain,pizzi2022self}.

Embeddings can be extracted from other modalities as well, with slight variations. 
For text embeddings (like BERT~\cite{devlin2018bert,izacard2021unsupervised}) the size of the embedding vector is usually \emph{larger} than the original text -- embeddings are not guaranteed to compress the items they are computed from. 
These text embeddings are the basis for the retrieval augmented generation for LLMs~\cite{izacard2020leveraging}. 
Text and image embeddings can be matched together as with the CLIP embeddings~\cite{radford2021learning}.
Recommendation systems are often based on two-tower networks, with different embedding functions for the users and the items to recommend~\cite{naumov2019deep}.

\section{Vector search with machine learning}
\label{sec:ML4VS}

Performing the embedding search of Equation~\ref{eq:embeddingsearch} is an operation whose complexity is $\mathcal{O}(N\times d)$ for most similarity functions. 
The research on vector search aims at reducing the runtime when $N$ becomes large, either by making it sub-linear or by reducing the linear factor. 
In the process, some accuracy can be sacrificed, and the brute force search of Equation~\ref{eq:embeddingsearch} becomes an Approximate Nearest Neighbor Search (ANNS) task.

\paragraph{Vector search is a linear classifier.}

Let's assume that the similarity function, in Equation~(\ref{eq:embeddingsearch}) is a dot product\footnote{This is without loss of generality: most standard distances can be reduced to computing dot products in a transformed space~\cite{douze2024faiss}.}. 
In that case, the operation to be performed is a matrix-vector multiplication between $E(q)$ and the matrix that stacks embeddings $[e_1,\dots,e_N]$. 
This is equivalent to a linear layer of size $N\times d$.
This means that vector search can be solved with the simplest of machine learning tools: a linear classifier of size $N$. 

This equivalence between vector search and classification is used for k-NN classifiers~\cite{mensink2013distance,caron2021emerging,qi2018low} (but k-NN classifiers are less accurate than training linear classifiers end-to-end). 

\paragraph{A fundamental divergence.}
ANNS and ML both strive to optimize an accuracy objective. 
However, vector search starts where machine learning stops.
The problems to handle when training a machine learning model are the train-test discrepancy, overfitting, regularization, etc.
Resource utilization is a second thought.
Vector search does not have any of these problems, because computing the perfect result is straightforward (Equation~\ref{eq:embeddingsearch}). 
The problem of vector search is how to compute the result accurately with limited resources. 

This explains why directly applying ML does not solve VS: as soon as ML starts to apply operations at scale $N$, it is already slower than performing burte-force VS.
This observation was already done in~\cite{kraska2018case}. 
Their approach is to break down the indexing structures into a distribution modeling part, that can be solved by ML and the indexing structure itself, that scales to size $N$.

\paragraph{ML for vector distribution modeling.}

A direct application of this approach for VS is to train a transformation that maps vectors to a space that is easier to index~\cite{sablayrolles2018spreading}.
In this approach the input vectors are transformed into a more uniform space while maintaining neighborhood relations; a lattice quantizer is then used to encode the uniform vectors. 

Modeling the vector distribution can be used to apply lossy compression to vectors, which is equivalent to quantization. 
The k-means algorithm is a strong baseline for quantization because it obeys Lloyd's conditions and is efficient.
In fact, the way VQ-VAEs are trained, with an exponential moving average, is similar to online k-means~\cite{razavi2019generating}.

However, to scale it to more accurate compression, k-means must be applied several times (multi-codebook quantization), which yields less optimal representations like product quantization~\cite{jegou2010product} or additive quantization~\cite{babenko2014additive}. 

Deep learning models can be applied to improve multi-codebook quantization. 
This is done in the UNQ quantizer~\cite{morozov2019unsupervised}, where a learned transformation maps the vectors to a space where it is easier to apply quantization. 
In the QINCo series of works~\cite{huijben2024QINco,vallaeys2025qinco2} the codebooks of a residual quantizer are adapted using a neural net to better approximate the vector distribution. 
One interesting work shows that, given a quantizer, it is possible to train a decoder that improves its accuracy~\cite{Amara2022NearestPerspective}.

One important family of VS methods is to partition the vector space. 
At search time, only a subset of partitions are visited. 
Usually this partitioning is based on k-means~\cite{jegou2010product}. 
However, the partitions can also be predicted from a vector with a classifier~\cite{dong2019learning,mazare2025inferencetimesparseattentionasymmetric}.

\paragraph{Discussion.}

These approaches show that there are operating points where ML can help improving VS. 
However, in many cases, the practitioner hits hard limits.
The first limit is that the runtime of the quantization must remain below that of a brute-force vector search. 
The second limit is that often, increasing the capacity of the model does not improve the fit of the vector distribution: 
a shallow model (or k-means itself) is hard to outperform with deeper models. 
Why this happens is an open research question.

\section{Machine learning with vector search}
\label{sec:VS4ML}

We review methods where VS is included within deep learning models. 

\paragraph{Network compression.}
In the previous section, we showed the equivalence between VS and a linear layer. 
Therefore, it is natural to apply vector search techniques to these layers, especially when the weight matrix is skinny ($N\gg d$).

Techniques originally developed for VS have been applied to compressing linear layers. 
Product quantization is used to compress convolutional neural nets~\cite{han2015deep} and language models~\cite{stock2019and}, with or without retraining the model (quantization-aware training or post-training quantization). 
This is especially useful for large recommendation models where most of the network parameters are in embedding tables~\cite{naumov2019deep}, \ie $N \gg d$.
In some settings, a compressed linear operator can be applied to its input without decompression~\cite{harchaoui2012large}.

\paragraph{Attention operators.}
Large language models are built around an attention operator that functions as an associative memory: a query vector is matched with embeddings of all the previous tokens of the prompt. 
This is basically a vector search task. 

When the prompt size increases (the ``long context'' setting, $N \gg d$), it becomes beneficial to use ANNS. 
This includes compression~\cite{egiazarian2024extreme}, and also non-exhaustive VS with partitions~\cite{bertsch2023unlimiformer}, random projections~\cite{zhuoming2024magicpig} or graphs~\cite{liu2024retrievalattention}.
One difficulty of attention operators is that the queries and database vectors have different distributions. 
For partition-based VS, this can be addressed by training different partition classifiers for queries and database vectors~\cite{mazare2025inferencetimesparseattentionasymmetric}.

\paragraph{Discussion.}
VS has a potential to be an accelerator for ML models, whenever the input needs to be compared to a large number of vectors.
The limit is that ML is typically run on hardware that is so efficient in doing brute-force VS (matrix multiplication)~\cite{chern2022tpu} that the size above which it makes sense to use ANNS is large (currently around $N=10^5$) and growing.

\section{Conclusion}

We reviewed several use cases where VS and ML are interlinked. 
It is clear that ML cannot replace the VS data structures, but it can help modeling the data distributions. 
One direction that deserves more exploration is leverage ML to guide the construction of graph-based indexes. 
In the other direction, there are tasks in ML where VS can be very useful, in particular to solve large classification problems and to perform attention on long contexts. 

\emph{The case for learned indexing structures}~\cite{kraska2018case} speculated that the increase of compute capacity would make ML models more amenable to indexing methods. 
What happened is rather that for many use cases of VS, compute has become so cheap that they can just be solved  in brute force.

\bibliographystyle{plain}
\bibliography{biblio}

\begin{thebibliography}{10}

\bibitem{Amara2022NearestPerspective}
Kenza Amara, Matthijs Douze, Alexandre Sablayrolles, and Hervé J{\'{e}}gou.
\newblock Nearest neighbor search with compact codes: A decoder perspective.
\newblock In {\em ICMR}, 2022.

\bibitem{aumuller2020ann}
Martin Aum{\"u}ller, Erik Bernhardsson, and Alexander Faithfull.
\newblock Ann-benchmarks: A benchmarking tool for approximate nearest neighbor
  algorithms.
\newblock {\em Information Systems}, 87:101374, 2020.

\bibitem{azizi2025graph}
Ilias Azizi, Karima Echihabi, and Themis Palpanas.
\newblock Graph-based vector search: An experimental evaluation of the
  state-of-the-art.
\newblock {\em arXiv preprint arXiv:2502.05575}, 2025.

\bibitem{babenko2014additive}
Artem Babenko and Victor Lempitsky.
\newblock Additive quantization for extreme vector compression.
\newblock In {\em CVPR}, 2014.

\bibitem{babenko2014neural}
Artem Babenko, Anton Slesarev, Alexandr Chigorin, and Victor Lempitsky.
\newblock Neural codes for image retrieval.
\newblock In {\em Computer Vision--ECCV 2014: 13th European Conference, Zurich,
  Switzerland, September 6-12, 2014, Proceedings, Part I 13}, pages 584--599.
  Springer, 2014.

\bibitem{bach2024learning}
Francis Bach.
\newblock {\em Learning theory from first principles}.
\newblock MIT press, 2024.

\bibitem{berman2019multigrain}
Maxim Berman, Herv{\'e} J{\'e}gou, Andrea Vedaldi, Iasonas Kokkinos, and
  Matthijs Douze.
\newblock Multigrain: a unified image embedding for classes and instances.
\newblock {\em arXiv preprint arXiv:1902.05509}, 2019.

\bibitem{bertsch2023unlimiformer}
Amanda Bertsch, Uri Alon, Graham Neubig, and Matthew~R Gormley.
\newblock Unlimiformer: Long-range transformers with unlimited length input.
\newblock {\em arXiv preprint arXiv:2305.01625}, 2023.

\bibitem{caron2018deep}
Mathilde Caron, Piotr Bojanowski, Armand Joulin, and Matthijs Douze.
\newblock Deep clustering for unsupervised learning of visual features.
\newblock In {\em Proceedings of the European conference on computer vision
  (ECCV)}, pages 132--149, 2018.

\bibitem{caron2021emerging}
Mathilde Caron, Hugo Touvron, Ishan Misra, Herv{\'e} J{\'e}gou, Julien Mairal,
  Piotr Bojanowski, and Armand Joulin.
\newblock Emerging properties in self-supervised vision transformers.
\newblock In {\em Proceedings of the IEEE/CVF international conference on
  computer vision}, pages 9650--9660, 2021.

\bibitem{chen2024roargraph}
Meng Chen, Kai Zhang, Zhenying He, Yinan Jing, and X~Sean Wang.
\newblock Roargraph: A projected bipartite graph for efficient cross-modal
  approximate nearest neighbor search.
\newblock {\em arXiv preprint arXiv:2408.08933}, 2024.

\bibitem{chen2020simple}
Ting Chen, Simon Kornblith, Mohammad Norouzi, and Geoffrey Hinton.
\newblock A simple framework for contrastive learning of visual
  representations.
\newblock In {\em International conference on machine learning}, pages
  1597--1607. PMLR, 2020.

\bibitem{zhuoming2024magicpig}
Zhuoming Chen, Ranajoy Sadhukhan, Zihao Ye, Jianyu Zhang, Niklas Nolte,
  Matthijs Douze, Leon Bottou, Zhihao Jia, and Beidi Chen.
\newblock Magicpig: Lsh sampling for efficient llm generation.
\newblock In {\em ArXiV}, 2024.

\bibitem{chern2022tpu}
Felix Chern, Blake Hechtman, Andy Davis, Ruiqi Guo, David Majnemer, and Sanjiv
  Kumar.
\newblock Tpu-knn: K nearest neighbor search at peak flop/s.
\newblock {\em Advances in Neural Information Processing Systems},
  35:15489--15501, 2022.

\bibitem{datar2004locality}
Mayur Datar, Nicole Immorlica, Piotr Indyk, and Vahab~S Mirrokni.
\newblock Locality-sensitive hashing scheme based on p-stable distributions.
\newblock In {\em Proceedings of the twentieth annual symposium on
  Computational geometry}, pages 253--262, 2004.

\bibitem{deng2019arcface}
Jiankang Deng, Jia Guo, Niannan Xue, and Stefanos Zafeiriou.
\newblock Arcface: Additive angular margin loss for deep face recognition.
\newblock In {\em Proceedings of the IEEE/CVF conference on computer vision and
  pattern recognition}, pages 4690--4699, 2019.

\bibitem{devlin2018bert}
Jacob Devlin, Ming-Wei Chang, Kenton Lee, and Kristina Toutanova.
\newblock Bert: Pre-training of deep bidirectional transformers for language
  understanding.
\newblock {\em arXiv preprint arXiv:1810.04805}, 2018.

\bibitem{dong2019learning}
Yihe Dong, Piotr Indyk, Ilya Razenshteyn, and Tal Wagner.
\newblock Learning space partitions for nearest neighbor search.
\newblock {\em arXiv preprint arXiv:1901.08544}, 2019.

\bibitem{dosovitskiy2021imageworth16x16words}
Alexey Dosovitskiy, Lucas Beyer, Alexander Kolesnikov, Dirk Weissenborn,
  Xiaohua Zhai, Thomas Unterthiner, Mostafa Dehghani, Matthias Minderer, Georg
  Heigold, Sylvain Gelly, Jakob Uszkoreit, and Neil Houlsby.
\newblock An image is worth 16x16 words: Transformers for image recognition at
  scale, 2021.

\bibitem{douze2024faiss}
Matthijs Douze, Alexandr Guzhva, Chengqi Deng, Jeff Johnson, Gergely Szilvasy,
  Pierre-Emmanuel Mazar{\'e}, Maria Lomeli, Lucas Hosseini, and Herv{\'e}
  J{\'e}gou.
\newblock The faiss library.
\newblock {\em arXiv preprint arXiv:2401.08281}, 2024.

\bibitem{egiazarian2024extreme}
Vage Egiazarian, Andrei Panferov, Denis Kuznedelev, Elias Frantar, Artem
  Babenko, and Dan Alistarh.
\newblock Extreme compression of large language models via additive
  quantization.
\newblock {\em arXiv preprint arXiv:2401.06118}, 2024.

\bibitem{gao2024rabitq}
Jianyang Gao and Cheng Long.
\newblock Rabitq: Quantizing high-dimensional vectors with a theoretical error
  bound for approximate nearest neighbor search.
\newblock {\em Proceedings of the ACM on Management of Data}, 2(3):1--27, 2024.

\bibitem{han2015deep}
Song Han, Huizi Mao, and William~J Dally.
\newblock Deep compression: Compressing deep neural networks with pruning,
  trained quantization and huffman coding.
\newblock {\em arXiv preprint arXiv:1510.00149}, 2015.

\bibitem{harchaoui2012large}
Zaid Harchaoui, Matthijs Douze, Mattis Paulin, Miroslav Dudik, and
  J{\'e}r{\^o}me Malick.
\newblock Large-scale image classification with trace-norm regularization.
\newblock In {\em 2012 IEEE conference on computer vision and pattern
  recognition}, pages 3386--3393. IEEE, 2012.

\bibitem{he2016deep}
Kaiming He, Xiangyu Zhang, Shaoqing Ren, and Jian Sun.
\newblock Deep residual learning for image recognition.
\newblock In {\em Proceedings of the IEEE conference on computer vision and
  pattern recognition}, pages 770--778, 2016.

\bibitem{huijben2024QINco}
Iris~A.M. Huijben, Matthijs Douze, Matthew~J. Muckley, Ruud~J.G. van Sloun, and
  Jakob Verbeek.
\newblock Residual quantization with implicit neural codebooks.
\newblock In {\em International Conference on Machine Learning (ICML)}, 2024.

\bibitem{izacard2021unsupervised}
Gautier Izacard, Mathilde Caron, Lucas Hosseini, Sebastian Riedel, Piotr
  Bojanowski, Armand Joulin, and Edouard Grave.
\newblock Unsupervised dense information retrieval with contrastive learning.
\newblock {\em arXiv preprint arXiv:2112.09118}, 2021.

\bibitem{izacard2020leveraging}
Gautier Izacard and Edouard Grave.
\newblock Leveraging passage retrieval with generative models for open domain
  question answering.
\newblock {\em arXiv preprint arXiv:2007.01282}, 2020.

\bibitem{jaiswal2020survey}
Ashish Jaiswal, Ashwin~Ramesh Babu, Mohammad~Zaki Zadeh, Debapriya Banerjee,
  and Fillia Makedon.
\newblock A survey on contrastive self-supervised learning.
\newblock {\em Technologies}, 9(1):2, 2020.

\bibitem{jegou2010product}
Herv\'e J\'egou, Matthijs Douze, and Cordelia Schmid.
\newblock Product quantization for nearest neighbor search.
\newblock {\em PAMI}, 2010.

\bibitem{jegou2010aggregating}
Herv{\'e} J{\'e}gou, Matthijs Douze, Cordelia Schmid, and Patrick P{\'e}rez.
\newblock Aggregating local descriptors into a compact image representation.
\newblock In {\em 2010 IEEE computer society conference on computer vision and
  pattern recognition}, pages 3304--3311. IEEE, 2010.

\bibitem{kraska2018case}
Tim Kraska, Alex Beutel, Ed~H Chi, Jeffrey Dean, and Neoklis Polyzotis.
\newblock The case for learned index structures.
\newblock In {\em Proceedings of the 2018 international conference on
  management of data}, pages 489--504, 2018.

\bibitem{liu2024retrievalattention}
Di~Liu, Meng Chen, Baotong Lu, Huiqiang Jiang, Zhenhua Han, Qianxi Zhang,
  Qi~Chen, Chengruidong Zhang, Bailu Ding, Kai Zhang, et~al.
\newblock Retrievalattention: Accelerating long-context llm inference via
  vector retrieval.
\newblock {\em arXiv preprint arXiv:2409.10516}, 2024.

\bibitem{malkov2018efficient}
Yu~A Malkov and Dmitry~A Yashunin.
\newblock Efficient and robust approximate nearest neighbor search using
  hierarchical navigable small world graphs.
\newblock {\em IEEE transactions on pattern analysis and machine intelligence},
  42(4):824--836, 2018.

\bibitem{mazare2025inferencetimesparseattentionasymmetric}
Pierre-Emmanuel Mazaré, Gergely Szilvasy, Maria Lomeli, Francisco Massa, Naila
  Murray, Hervé Jégou, and Matthijs Douze.
\newblock Inference-time sparse attention with asymmetric indexing, 2025.

\bibitem{mensink2013distance}
Thomas Mensink, Jakob Verbeek, Florent Perronnin, and Gabriela Csurka.
\newblock Distance-based image classification: Generalizing to new classes at
  near-zero cost.
\newblock {\em IEEE transactions on pattern analysis and machine intelligence},
  35(11):2624--2637, 2013.

\bibitem{morozov2019unsupervised}
Stanislav Morozov and Artem Babenko.
\newblock Unsupervised neural quantization for compressed-domain similarity
  search.
\newblock In {\em ICCV}, 2019.

\bibitem{naumov2019deep}
Maxim Naumov, Dheevatsa Mudigere, Hao-Jun~Michael Shi, Jianyu Huang, Narayanan
  Sundaraman, Jongsoo Park, Xiaodong Wang, Udit Gupta, Carole-Jean Wu,
  Alisson~G Azzolini, et~al.
\newblock Deep learning recommendation model for personalization and
  recommendation systems.
\newblock {\em arXiv preprint arXiv:1906.00091}, 2019.

\bibitem{oquab2024dinov2learningrobustvisual}
Maxime Oquab, Timothée Darcet, Théo Moutakanni, Huy Vo, Marc Szafraniec,
  Vasil Khalidov, Pierre Fernandez, Daniel Haziza, Francisco Massa, Alaaeldin
  El-Nouby, Mahmoud Assran, Nicolas Ballas, Wojciech Galuba, Russell Howes,
  Po-Yao Huang, Shang-Wen Li, Ishan Misra, Michael Rabbat, Vasu Sharma, Gabriel
  Synnaeve, Hu~Xu, Hervé Jegou, Julien Mairal, Patrick Labatut, Armand Joulin,
  and Piotr Bojanowski.
\newblock Dinov2: Learning robust visual features without supervision, 2024.

\bibitem{perronnin2010large}
Florent Perronnin, Yan Liu, Jorge S{\'a}nchez, and Herv{\'e} Poirier.
\newblock Large-scale image retrieval with compressed fisher vectors.
\newblock In {\em 2010 IEEE computer society conference on computer vision and
  pattern recognition}, pages 3384--3391. IEEE, 2010.

\bibitem{pizzi2022self}
Ed~Pizzi, Sreya~Dutta Roy, Sugosh~Nagavara Ravindra, Priya Goyal, and Matthijs
  Douze.
\newblock A self-supervised descriptor for image copy detection.
\newblock In {\em Proceedings of the IEEE/CVF Conference on Computer Vision and
  Pattern Recognition}, pages 14532--14542, 2022.

\bibitem{qi2018low}
Hang Qi, Matthew Brown, and David~G Lowe.
\newblock Low-shot learning with imprinted weights.
\newblock In {\em Proceedings of the IEEE conference on computer vision and
  pattern recognition}, pages 5822--5830, 2018.

\bibitem{radford2021learning}
Alec Radford, Jong~Wook Kim, Chris Hallacy, Aditya Ramesh, Gabriel Goh,
  Sandhini Agarwal, Girish Sastry, Amanda Askell, Pamela Mishkin, Jack Clark,
  et~al.
\newblock Learning transferable visual models from natural language
  supervision.
\newblock In {\em International conference on machine learning}, pages
  8748--8763. PMLR, 2021.

\bibitem{razavi2019generating}
Ali Razavi, Aaron Van~den Oord, and Oriol Vinyals.
\newblock Generating diverse high-fidelity images with vq-vae-2.
\newblock {\em Advances in neural information processing systems}, 32, 2019.

\bibitem{sablayrolles2018spreading}
Alexandre Sablayrolles, Matthijs Douze, Cordelia Schmid, and Herv{\'e}
  J{\'e}gou.
\newblock Spreading vectors for similarity search.
\newblock {\em ICLR}, 2019.

\bibitem{bigann23}
Harsha~Vardhan Simhadri, Martin Aum{\"u}ller, Amir Ingber, Matthijs Douze,
  George Williams, Magdalen~Dobson Manohar, Dmitry Baranchuk, Edo Liberty,
  Frank Liu, Ben Landrum, et~al.
\newblock Results of the big ann: Neurips'23 competition.
\newblock {\em arXiv preprint arXiv:2409.17424}, 2024.

\bibitem{bigann21}
Harsha~Vardhan Simhadri, George Williams, Martin Aum{\"u}ller, Matthijs Douze,
  Artem Babenko, Dmitry Baranchuk, Qi~Chen, Lucas Hosseini, Ravishankar
  Krishnaswamny, Gopal Srinivasa, et~al.
\newblock Results of the neurips’21 challenge on billion-scale approximate
  nearest neighbor search.
\newblock In {\em NeurIPS 2021 Competitions and Demonstrations Track}, pages
  177--189. PMLR, 2022.

\bibitem{stock2019and}
Pierre Stock, Armand Joulin, R{\'e}mi Gribonval, Benjamin Graham, and Herv{\'e}
  J{\'e}gou.
\newblock And the bit goes down: Revisiting the quantization of neural
  networks.
\newblock {\em arXiv preprint arXiv:1907.05686}, 2019.

\bibitem{subramanya2019diskann}
Suhas~Jayaram Subramanya, Rohan Kadekodi, Ravishankar Krishaswamy, and
  Harsha~Vardhan Simhadri.
\newblock Diskann: Fast accurate billion-point nearest neighbor search on a
  single node.
\newblock In {\em NeurIPS}, 2019.

\bibitem{vallaeys2025qinco2}
Th{\'e}ophane Vallaeys, Matthew Muckley, Jakob Verbeek, and Matthijs Douze.
\newblock Qinco2: Vector compression and search with improved implicit neural
  codebooks.
\newblock In {\em ICLR}, 2025.

\end{thebibliography}

\end{document}